\documentclass{article}

\usepackage{microtype}
\usepackage{graphicx}
\usepackage{subfigure}
\usepackage{tikz}
\usepackage{booktabs} 
\usepackage{amsmath}

\usepackage{hyperref}

\usepackage[accepted]{icml2018}

\icmltitlerunning{Behavior Learning with STDP, Memory, and Intrinsic Reward}

\begin{document}

\twocolumn[
\icmltitle{Learning as Reinforcement: Applying Principles of Neuroscience for More General Reinforcement Learning Agents}




\begin{icmlauthorlist}
\icmlauthor{Eric Zelikman}{stansym}
\icmlauthor{William Yin}{stan}
\icmlauthor{Kenneth Wang}{stan}
\end{icmlauthorlist}

\icmlaffiliation{stansym}{Department of Symbolic Systems, Stanford University, California, USA}
\icmlaffiliation{stan}{Department of Computer Science, Stanford University, California, USA}

\icmlcorrespondingauthor{Kenneth Wang}{kwang411@stanford.edu}
\icmlcorrespondingauthor{William Yin}{wyin@stanford.edu}
\icmlcorrespondingauthor{Eric Zelikman}{ezelikman@stanford.edu}

\icmlkeywords{Learning as Reinforcement: Applying Principles of Neuroscience for More General Reinforcement Learning Agents}

\vskip 0.3in
]



\printAffiliationsAndNotice{}  

\begin{abstract}
A significant challenge in developing AI that can generalize well is designing agents that learn about their world without being told what to learn, and apply that learning to challenges with sparse rewards. Moreover, most traditional reinforcement learning approaches explicitly separate learning and decision making in a way that does not correspond to biological learning. We implement an architecture founded in principles of experimental neuroscience, by combining computationally efficient abstractions of biological algorithms. Our approach is inspired by research on spike-timing dependent plasticity, the transition between short and long term memory, and the role of various neurotransmitters in rewarding curiosity. The Neurons-in-a-Box architecture can learn in a wholly generalizable manner, and demonstrates an efficient way to build and apply representations without explicitly optimizing over a set of criteria or actions. We find it performs well in many environments including OpenAI Gym's Mountain Car, which has no reward besides touching a hard-to-reach flag on a hill, Inverted Pendulum, where it learns simple strategies to improve the time it holds a pendulum up, a video stream, where it spontaneously learns to distinguish an open and closed hand, as well as other environments like Google Chrome's Dinosaur Game. 
\end{abstract}

\section{Introduction}
\label{sec:intro}
Previous works introduce the concept of spiking neural networks and illustrate reinforcement learning as a potential use case \cite{DONAHOE1997336, Stanley:2002:ERL:2955491.2955578, 1595864}. We build upon these works by abstracting away the necessity for spike trains and introducing new ideas from neuroscience, including memory formation, intrinsic motivation, and computationally-efficient algorithms for spike-timing dependent plasticity.

\begin{figure}[h!]
\centering
\includegraphics[scale=0.6]{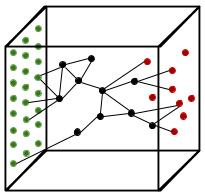}
\caption{Neurons-in-a-Box architecture. Green represents input neurons (i.e. pixel information), while red neurons are randomly selected from a slice of the box. Black neurons are the intermediary neurons (of which only a subset is displayed).}
\end{figure}

\section{Background}
\subsection{Spike-Timing Dependent Plasticity}
Spike-timing dependent plasticity (STDP) is a biological phenomenon that constantly updates the strength of connections between neurons in the brain based on firing patterns. Connections strengths are adjusted based on the timings between a neuron firing and its input neurons firing, such that neurons that fire consecutively have their connections strengthened in the ``direction'' of firing. This formulation has been shown to explain characteristics of long-term depression and other neuroscientific phenomena, as shown in \cite{debanne1994asynchronous}.

One natural but generally not-revealing approach is the use of STDP to train a neural network to identify patterns, then using a separate neural network trained on the STDP network's firings to make predictions. This approach was used to accurately predict road traffic dynamics by \citet{kasabov2014neucube}.

Another important adaptation of STDP to the formulation of reinforcement learning is called reward-modulated STDP, as explained in \cite{florian2007reinforcement}. The idea is that including a global reward signal into the standard STDP formulation can be analytically shown to lead to a reinforcement learning algorithm, thus ``modulating'' the STDP with the reward. In essence, this is simply STDP where the learning update step is multiplied by the reward. We see that in a later section, we continue to draw upon neuroscience research by coming up with intrinsic rewards based on motivation rather than explicit rewards specified by the environment or programmer.

\subsection{Long-Term and Short-Term Memory}
Research indicates that there are distinct chemical and physiological properties of neural connections which correspond to long-term and short-term memory \cite{pmid25007209}. Moreover, there seem to be explicit chemical causes for this transition \cite{10.3389/fncir.2013.00052}. In the context of spike-timing dependent plasticity, a transition from short-term to long-term memory of a connection that has been consistently, actively predictive allows for more sophisticated new representations to be built ones. 

\subsection{Intrinsic Reward}
Intrinsic reward refers to the idea that humans often learn and discover useful behaviors within their environments through experimentation, exploration, and often just spontaneity. In work modeling intrinsic reward signals in artificial agents, researchers frequently draw an analogy with young infants, who are often observed to learn about the world by engaging with it through "playful behavior" \cite{2018arXiv180207442H}. The goal in similar research is discovering a reward signal which allows an agent to learn effectively about its environment in a generalizable manner, regardless of the structure of the environment. 

\subsection{Neural Noise}
Some amount of neural noise has consistently been shown to be useful in training spiking neural models and appears to be important in actual neuronal processing, though aside from possibly preventing overfitting, the function this noise plays in the actual brain is a topic of debate \cite{Longtin:2013, Engel2015}. Regardless, some baseline level of firing encourages a model that does something rather than nothing.

\section{Approach}
\label{sec:approach}

\subsection{Architecture}
As our model, we constructed a Neuron-in-a-Box architecture. This architecture was composed of an input layer which took in pixel input (either from a continuous video and/or audio stream) which then produced feedforward outputs into a number of feedforward neurons. These neurons were randomly spatially distributed, and connections were formed such that neurons closer spatially would be more likely to form a connection. Then, a random subset of neurons in a slice of the architecture are selected as output neurons, which were moved to a constant distance from the input. 

The long-term memory was implemented by keeping an accumulator which tracks the frequency with which a neuron fires and, if synaptic strength is high enough, changes the plasticity of the connection to false. 

The intrinsic reward which we attempted to use was simply the amount of learning that occurs at each timestep. In other words, we expected that training based on amount of novelty would eventually lead to good performance in a similar manner to a curious agent exploring phenomena it finds surprising. Unfortunately, in the dinosaur game, the game-over screen results in more novelty than simply continuing to explore the world, so eventually the agent converges on immediately ending the game. While in theory the game-over screen should eventually become uninteresting to the agent, this did not seem to happen in practice. 

\subsection{Reinforcement}
One guiding principle used was the idea that, at least at this level, a small number of neurons do not explicitly learn state values and choose firings that maximize over these states. Instead, the presence or lack of reward should be used to directly strengthen or weaken neural connections.

\subsubsection{Direct Reward}

Suppose we take a very blunt approach: if the neural network did something good, we want it to do that more often. If it did something bad, we want it to do less of that. This "global" update seems pretty consistent with what we see in the brain: many reward pathways, when triggered, affect many neurons at the same time and propagate to large parts of the brain \cite{wise1989brain}.  

There are several questions that need to be answered with this approach: first, what should be strengthened with a global reinforcement? For consistency, I will refer to the underlying sensory inputs as sensory neurons, the neurons that are not directly measured when determining an action as intermediate neurons, and the neurons that are directly measured as action neurons.

\begin{enumerate}
    \item \textbf{All neural connections (including inhibitory connections)}: This is far too coarse, and actually weakens the strongest connections relative to everything else due to the maximal connection strength. It introduces noise and encourages unstable solutions. However, combined with STDP, this can encourages the learning of meaningful relationships because neurons that may have nearly fired before, that capture information relevant to the reward, will now be more likely to fire and be related to one another by an action neuron.
    \item \textbf{Outputs from neurons that fired}: This is more stable, but has another significant problem: sometimes you want to reward passivity. For example, in the pendulum game, the action neurons are rewarded for not firing when the pendulum is at the top. Furthermore, as we know from backpropagation, which we avoided using, not the entire network is equally responsible for the result. 
    \item \textbf{Connections which were used}: That is, where the input neuron fired and the output neuron's firing corresponds to whether the connection is excitatory (positive) or inhibitory (negative).  This is one of the better options, but it is almost always more consequential to fire than to not fire so this ends up killing the network.
    \item \textbf{Input connections to action neurons}: This is stable and can actually indirectly facilitate the development of "reward neurons," but can actually result in a disproportionate reinforcement of inputs which are counterproductive (for example, if an action neuron fired with to one strong excitatory input and several weak inhibitory inputs, it would actually become less likely to fire). If instead you consider only the connections that were used, as in the previous item, one could end up silencing all connections there's a delay between reward and firing.
    \item \textbf{Input connections to action neurons where both fired}: This is stable and results in behavior learning, but unfortunately still suffers from the same problem as Approach 2. However, STDP can accomplish much of the work which backpropogation is usually used for, allowing neurons that are strengthened to strengthen their inputs which are responsible.
\end{enumerate}

\subsubsection{STDP Reinforcement}
There are two ways that we considered using STDP to directly shape behavior. First, we can simply multiply the STDP reward update by the reward. For negative rewards, this will actually lead to unlearning or "forgetting" whatever was just experienced, which isn't necessarily ideal. Alternatively, we can measure the size of the learning update and using a direct reward to reinforce actions that lead to more learning. This is necessary to approach questions with sparse rewards that require exploration. In essence, it encourages the network to take a more scenic route when it can. 

\subsection{Memory}
The idea of memory in our model is to keep fixed those connections that lead to the agent's favorable behavior so that intelligent strategies are propagated forward in time. Our approach consists of a plasticity mask which modulates (in the short term) the degree with which any connection can be altered by future weight updates in the long term. To implement this in practice, we considered two approaches.

\begin{enumerate}
    \item \textbf{All neural connections decay uniformly ("aging"):} This method involves decaying the plasticity of all connections uniformly over time. This approach seems far too coarse as it fails to adequately account for differences in favorability of connections.
    \item \textbf{Plasticity decay accumulation:} This approach involves accumulating modifications to individual connections. In other words, the more a particular connection is changed, the more 'change' is accumulated. Once this value reaches a certain threshold for a connection and its strength is significantly above mean connection strength, the connection is fixed and cannot be altered. This relies on the assumption that those connections which have been altered many times are closer to converging on an optimal fixed strength, whereas those which have not changed much still have room to change.
\end{enumerate}

\section{Theory}
There are a number of important theoretical questions which needed to be tackled in order to implement this agent.

\subsection{Discrete STDP Update Rule}
STDP is often implemented based on a weighted sum over some rolling windows combined with some neuronal spiking frequency \cite{10.3389/fnins.2018.00435}. This approach is absolutely biologically realistic, and researchers have even argued that norepinephrine plays the role of determining the size of this window \cite{Salgado2012}. However, it's computationally inefficient to implement across many neurons, and is only parallelizable with a high STDP update frequency, which comes with its own computational cost. 

We derive a discrete, matrix-operation parallel to the standard STDP rule. Given with $n$ neurons, a prior time-window of firings $\alpha \in \{0, 1\}^{n}$ and a current time-window firings $\beta \in \{0, 1\}^{n}$, an an input connection matrix C $\in \mathbf{R}^{n x n}$ we want to output an updated connection matrix C' $\in \mathbf{R}^{n x n}$. 

We want a function with two features. 

\begin{itemize}
    \item It should strengthen connections which are predictive: if neuron $A$ fires and then neuron $B$ fires, the connection $AB$ should strengthen and $BA$ should weaken.
    \item It should only strengthen connections where both of the neurons fired at least once
\end{itemize}

\subsubsection{Predictivity}
We can represent the first feature by the following rule: For first firings $\alpha$ and second firing $\beta$ as defined above,
\begin{equation}
    f(\alpha,\beta) = \beta - (1 - \alpha^T).
\end{equation}

For $\alpha = (1,0)$,
\begin{equation}
\beta=(0,1) \rightarrow f_0(\alpha,\beta) =
\begin{bmatrix} 
0 & 1 \\
-1 & 0 
\end{bmatrix}.
\end{equation}

For $\alpha = (1,1)$, 
\begin{equation}
\beta=(0,1) \rightarrow f_0(\alpha,\beta) =
\begin{bmatrix} 
0 & 1 \\
0 & 1 
\end{bmatrix}.
\end{equation}

For $\alpha = (0,0)$, 
\begin{equation}
\beta=(0,1) \rightarrow f_0(\alpha,\beta) =
\begin{bmatrix} 
-1 & 0 \\
-1 & 0 
\end{bmatrix}.
\end{equation}

Since the last example is clearly not the desired result (One doesn't want to weaken all connections that fire unprompted), we need to incorporate the second requirement.

\subsubsection{Co-occurrence}
To do this, we calculate the pairwise matrix directly: \begin{equation}
f_1(\alpha,\beta) = (\beta \alpha^T) | (\alpha \beta^T).
\end{equation} 

For example, for $\alpha = (0,0, 1)$,
\begin{equation*}
    \beta=(1,0,1) \rightarrow f_1(\alpha,\beta) =
\end{equation*}
\begin{equation}
\begin{bmatrix} 
0 & -1 & 0 \\
0 & -1 & 0 \\
1 & 0 & 1 \\
\end{bmatrix} * 
\begin{bmatrix} 
0 & 0 & 1 \\
0 & 0 & 0 \\
1 & 0 & 1 \\
\end{bmatrix}
= 
\begin{bmatrix} 
0 & 0 & 0 \\
0 & 0 & 0 \\
1 & 0 & 1 \\
\end{bmatrix}
\end{equation}

That is, the connection from the last neuron to the first neuron is the only one that's strengthened. 

\subsubsection{Update}
Supposing we have a connection matrix, C, of 
\begin{equation*}
\begin{bmatrix} 
0 & a_b & a_c \\
b_a & 0 & b_c \\
c_a & c_b & 0 \\
\end{bmatrix},
\end{equation*}

an update weight of $\gamma$, and a plasticity matrix $P$ of all 1's. The new connection matrix is
\begin{equation}
C' = f(\alpha, \beta, C) =
\end{equation}
\begin{equation*}
C + \gamma P \begin{bmatrix} 
0 & a_b & a_c \\
b_a & 0 & b_c \\
c_a & c_b & 0 \\
\end{bmatrix} * \begin{bmatrix} 
0 & 0 & 0 \\
0 & 0 & 0 \\
1 & 0 & 1 \\
\end{bmatrix} = 
\begin{bmatrix} 
0 & a_b & a_c \\
b_a & 0 & b_c \\
c_a + \gamma & c_b & 0 \\
\end{bmatrix}.
\end{equation*}

\subsubsection{Overrall}
Where for the firings at to timesteps, $\alpha,\beta$, a learning rate $\gamma$, a connections matrix C, and a plasticity matrix P, the discrete STDP update rule is:

$$f_0(\alpha,\beta) = \beta - (1 - \alpha^T)$$
$$f_1(\alpha,\beta) = (\beta \alpha^T) | (\alpha \beta^T)$$
$$C' = C + \gamma P \circ f_0(\alpha, \beta) \circ  f_1(\alpha, \beta)$$

\subsection{Reward Neurons}

Reward neurons are a surprisingly elegant way to learn rewards and modulate actions. Essentially, if one sets the reward neuron to fire with strength corresponding the reward, then an STDP update will reinforce other input connections that predict the firing of that reward, and then inputs that predict the firings of those inputs will be further strengthened. 

\begin{center}
\begin{tikzpicture}[scale=0.15]
\tikzstyle{every node}+=[inner sep=0pt]
\draw [black] (38,-11.9) circle (3);
\draw (38,-11.9) node {$r$};
\draw [black] (50.8,-11.9) circle (3);
\draw (50.8,-11.9) node {$R$};
\draw [black] (50.8,-11.9) circle (2.4);
\draw [black] (38,-21.5) circle (3);
\draw (38,-21.5) node {$B$};
\draw [black] (38,-31.7) circle (3);
\draw (38,-31.7) node {$A$};
\draw [black] (28.3,-31.7) circle (3);
\draw (28.3,-31.7) node {$\beta_1$};
\draw [black] (48.3,-31.7) circle (3);
\draw (48.3,-31.7) node {$\beta_2$};
\draw [black] (47.8,-11.9) -- (41,-11.9);
\fill [black] (41,-11.9) -- (41.8,-12.4) -- (41.8,-11.4);
\draw (44.4,-12.4) node [below] {$=$};
\draw [black] (38,-28.7) -- (38,-24.5);
\fill [black] (38,-24.5) -- (37.5,-25.3) -- (38.5,-25.3);
\draw (37.5,-26.6) node [left] {$c_{AB}$};
\draw [black] (38,-18.5) -- (38,-14.9);
\fill [black] (38,-14.9) -- (37.5,-15.7) -- (38.5,-15.7);
\draw (38.5,-16.7) node [right] {$c_{Br}$};
\draw [black] (30.37,-29.53) -- (35.93,-23.67);
\fill [black] (35.93,-23.67) -- (35.02,-23.91) -- (35.74,-24.6);
\draw [black] (46.17,-29.59) -- (40.13,-23.61);
\fill [black] (40.13,-23.61) -- (40.35,-24.53) -- (41.05,-23.82);
\end{tikzpicture}
\end{center}

Ideally, we want B to also be able to trigger the reward neuron, so that reward can be predicted when not yet present, so the reward neuron should sum its inputs, rather than simply be set directly to R. Since $c_{Br}$ is initially weak, then this is initially $R$. 

If r and B have a firing covariance $Cov(B, r)$, eventually $c_{Br}$ will converge on $Cov(B, r)$, if there are no other outputs, since $c_{Br}$ will be strengthened if the frequency of firing of B is less than the firing of $r$, and weakened otherwise. If B is an output neuron, then this implies a statistical equivalent to Rescorla-Wagner learning \cite{rescorla1972theory}\footnote{Where V is the current value of a state, $\alpha$ is the learning rate}: 
\begin{equation}
V' = V + \alpha (r - V).
\end{equation}
The more similar the frequencies of firing, the larger the update from an error in the opposite direction (i.e. the closer it gets to the actual covariance) and the smaller an error in the correct direction. 

Note that this same relationship between also applies between A and B, so $Cov(A, B) * Cov(B, r)$ tends to $Cov(A, r)$. As a result, as the connection between $B$ and $r$ is strengthened in line with Rescorla-Wagner, the same thing happens to the connection from $A$ to $B$. Propagating R-W value backwards in time by considering the covariance of states in this way is known as Temporal Difference learning: 
\begin{equation}
V'(s) = V(s)+\alpha (r+\gamma V(s')-V(s)).
\end{equation}

\subsection{Novelty}
Novelty as a reward has been shown for decades in primates to be a valid reward stimulus\cite{Martinez-Rubio1942}. Hence, we aimed to model novelty as reward in environments in which a clear, continuous reward signal is not present. 

As defined earlier, $f(\alpha, \beta)$ is the STDP update matrix of firings $\alpha$ and $\beta$. (In practice, these correspond to the last firings and the second-to-last firings.) $C$ is our last connection matrix, $P$ is our plasticity matrix, $N$ is our novelty, and $n$ is the size (i.e. number of elements) of $C$. Thus, we have
\begin{equation}
N = \overline{M},
\end{equation}
where 
\begin{equation}
\overline{M} = \frac{1}{n} \sum_{i=1}^n |f(\alpha, \beta) * C * P|.
\end{equation}
In other words, we are taking the mean over the magnitude of change between firing updates. 

An alternative measure of novelty would be performing this same operation using video frames rather than firing patterns (i.e. $M = |\alpha - \beta|$, where $\alpha$ and $\beta$ are the last and second-to-last video frames). This measure of novelty was used in testing on the Dino game environment, described below.

\section{Experimental Results}
\subsection{Pattern Recognition and Differentiation}

\begin{figure}
\centering
\includegraphics[width=0.4\columnwidth]{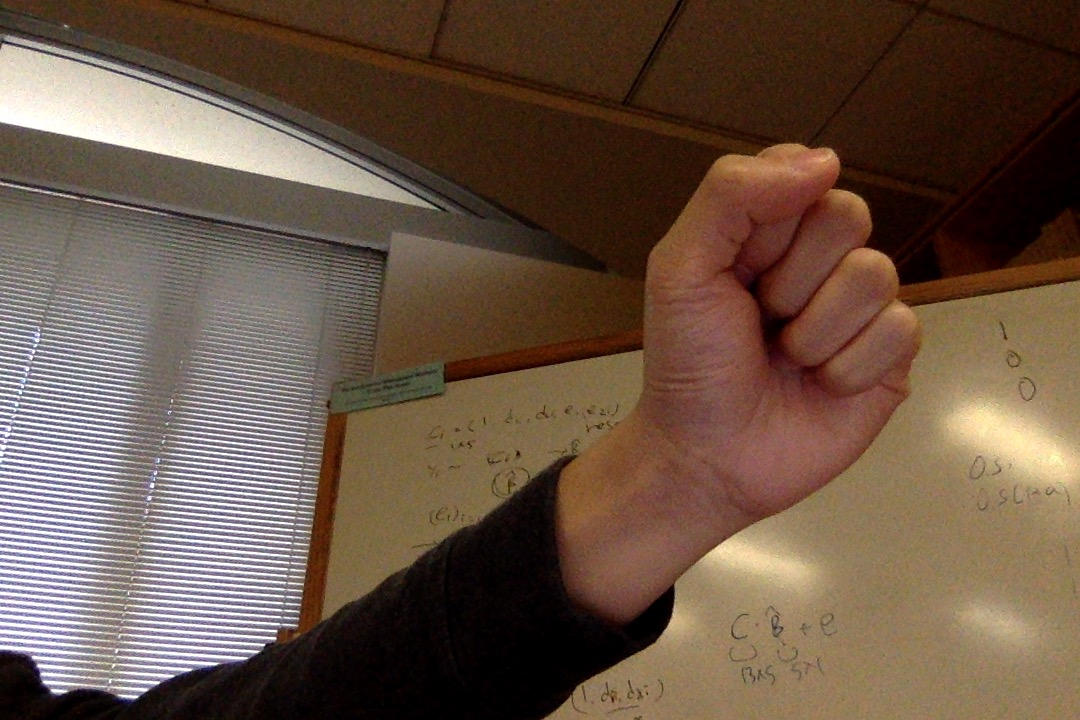}
\includegraphics[width=0.4\columnwidth]{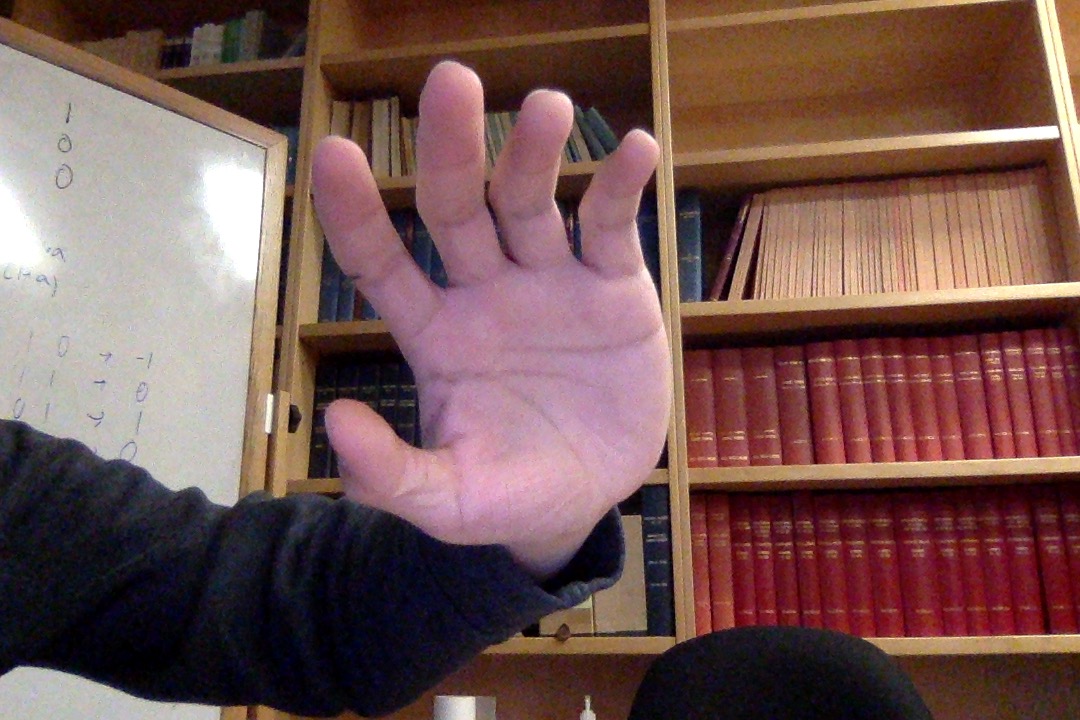}
\caption{
\label{fig:handpics}
Example visual stimuli shown to the network through a camera video stream. These were two differentiable stages in one particular experiment, composed of a closed fist and an open palm. Notice that the agent has no issue differentiating between the two stimuli, despite both the hand and the background being arbitrarily positioned.}
\end{figure}

As an initial experiment, STDP was applied to neurons at each timestep which had, as input, a video stream. The neural network had no modulation of learning, and were essentially driven to build correlations between the data presented, both spatially and temporally. In this way, the network learned to differentiate between patterns observed in the video stream. To evaluate the model's effectiveness, a hand was shown to the neurons in different stages (e.g. in a fist or open palm, as shown in Figure~\ref{fig:handpics}) in different locations on the screen. 

\begin{figure}
\centering
\includegraphics[width=0.7
\columnwidth]{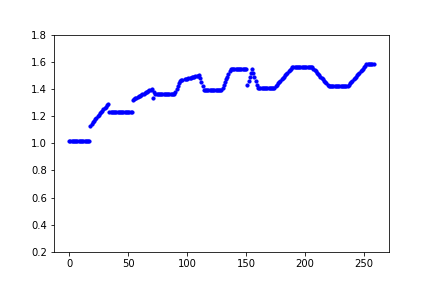}
\caption{ 
\label{fig:handgraph}
Ratio of differences in firing between the two neurons which best predicted which stage was currently occurring, one shown gestures corresponding to the stage and one not. The peaks and valleys correspond to alternations of stages.}
\end{figure}

The stages alternated at varying times and the neurons were recorded to keep track the maximal difference between any neuron's tendency to fire in the first and second stages. For a control, this process was run again where nothing was changed between stages. The neuron with the most variance between stages is chosen as the output neuron. Then, the ratio of the differences in activation of this neuron in the control and trained cases is shown in Figure~\ref{fig:handgraph}.

\begin{figure}
\centering
\includegraphics[width=0.2 \columnwidth]{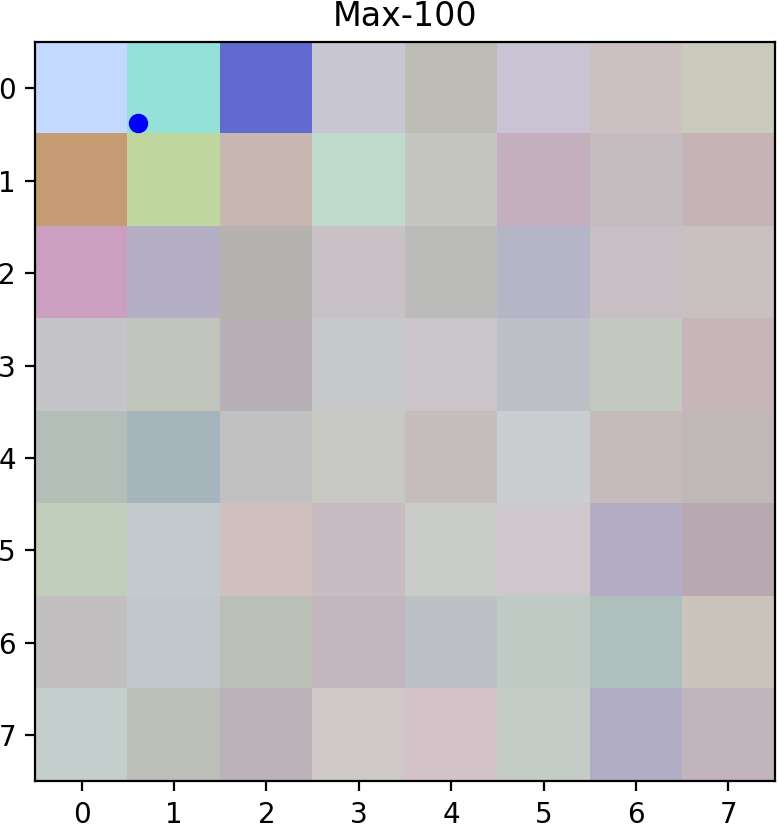}
\includegraphics[width=0.2 \columnwidth]{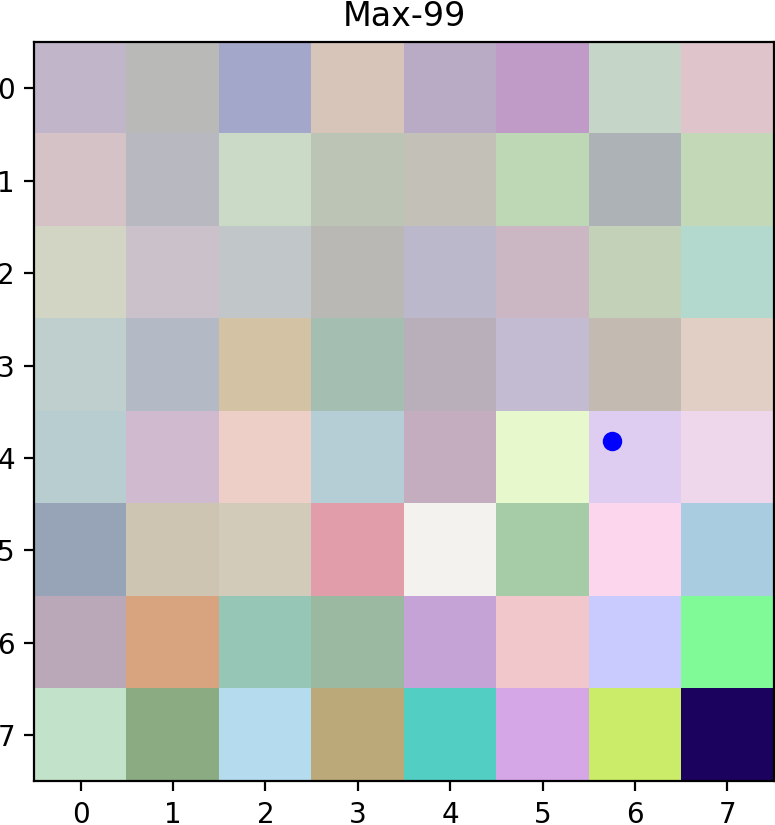}
\includegraphics[width=0.2 \columnwidth]{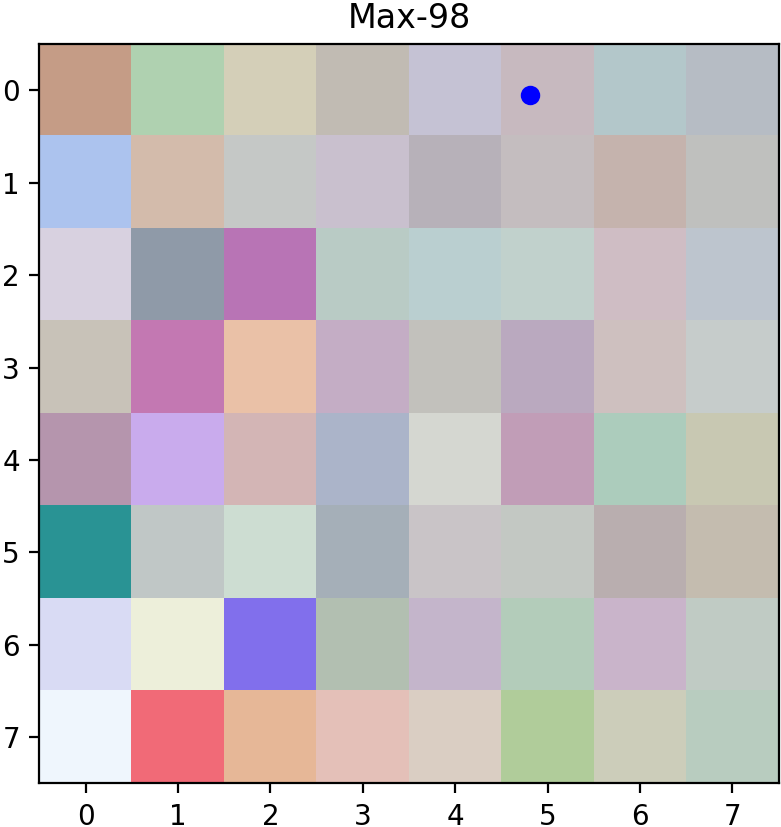}

\includegraphics[width=0.2 \columnwidth]{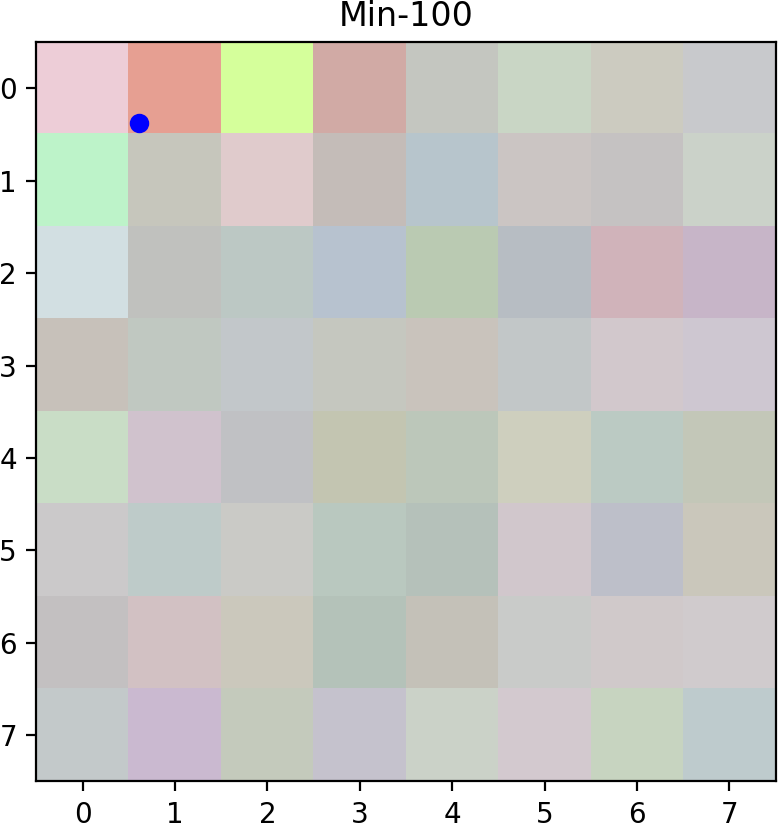}
\includegraphics[width=0.2 \columnwidth]{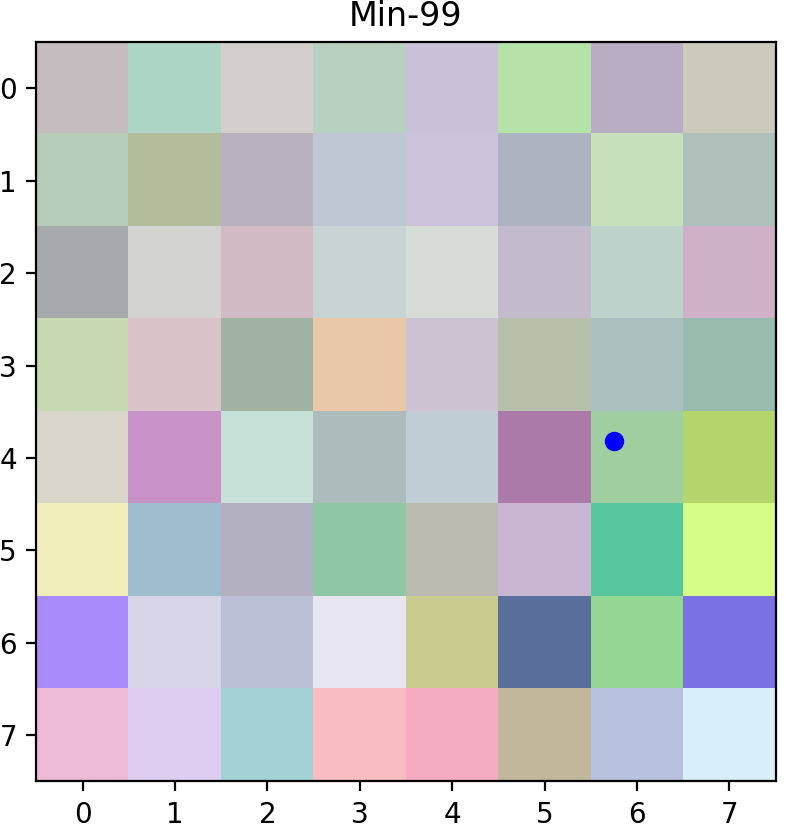}
\includegraphics[width=0.2 \columnwidth]{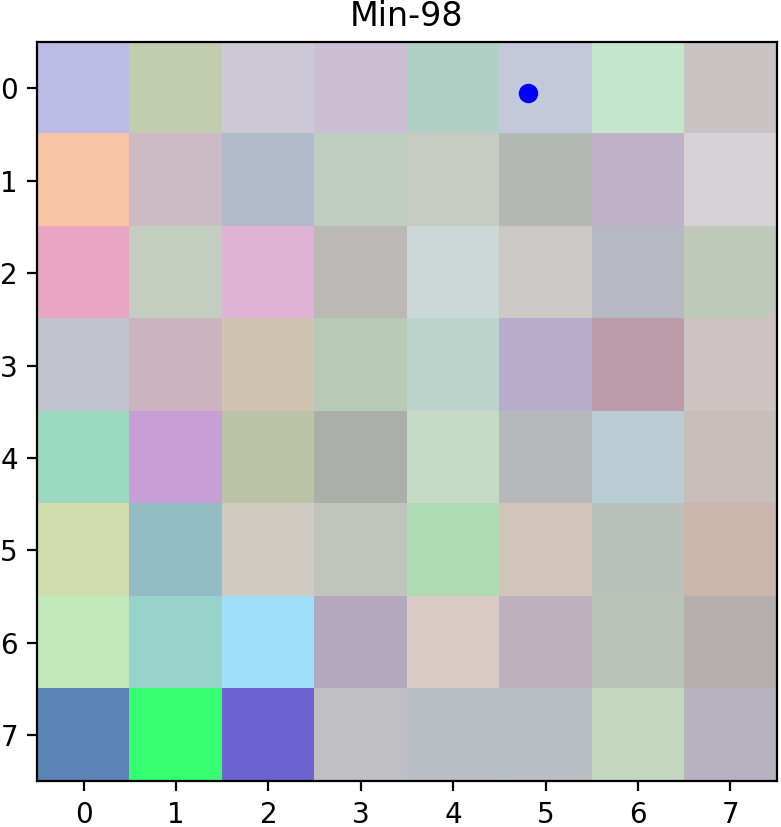}

\caption{
\label{fig:pixels}
Visualization of the images that consistently stimulated various neurons which had watched a few hours of low-resolution YouTube videos. The top row corresponds to the inputs that cause as much stimulation as possible in three neurons, while the bottom row corresponds to inputs that cause as little stimulation as possible in the same neurons. The three neurons each became sensitive (or insensitive) to a particular region of the input space.}
\end{figure}

This network was also shown YouTube for several hours, and then we visualized visual stimuli for the neurons that were most sensitive. Specifically, the neurons were activated by a large number\footnote{Typically, $n=1000$} of samples of random noise, and the sample that activated them most was selected, and then averaged again with many samples of random noise, with this process repeated several times. This whole process was repeated several times and the images which were stimulating or inhibitory were averaged together. We present these results in Figure~\ref{fig:pixels}.

\subsection{XOR}

The XOR gate problem has historically been known to be notoriously difficult to learn due to its nonlinear structure.\footnote{In what was known as the XOR affair, Marvin Minsky famously mathematically proved that a single-layer perceptron was incapable of solving the XOR gate. This was among many reasons for the AI winter of the 1970s.} The problem involves returning true only when one of two bits have fired, and false otherwise. 

\begin{figure}
\centering
\includegraphics[width=0.8\columnwidth, height=0.5\columnwidth]{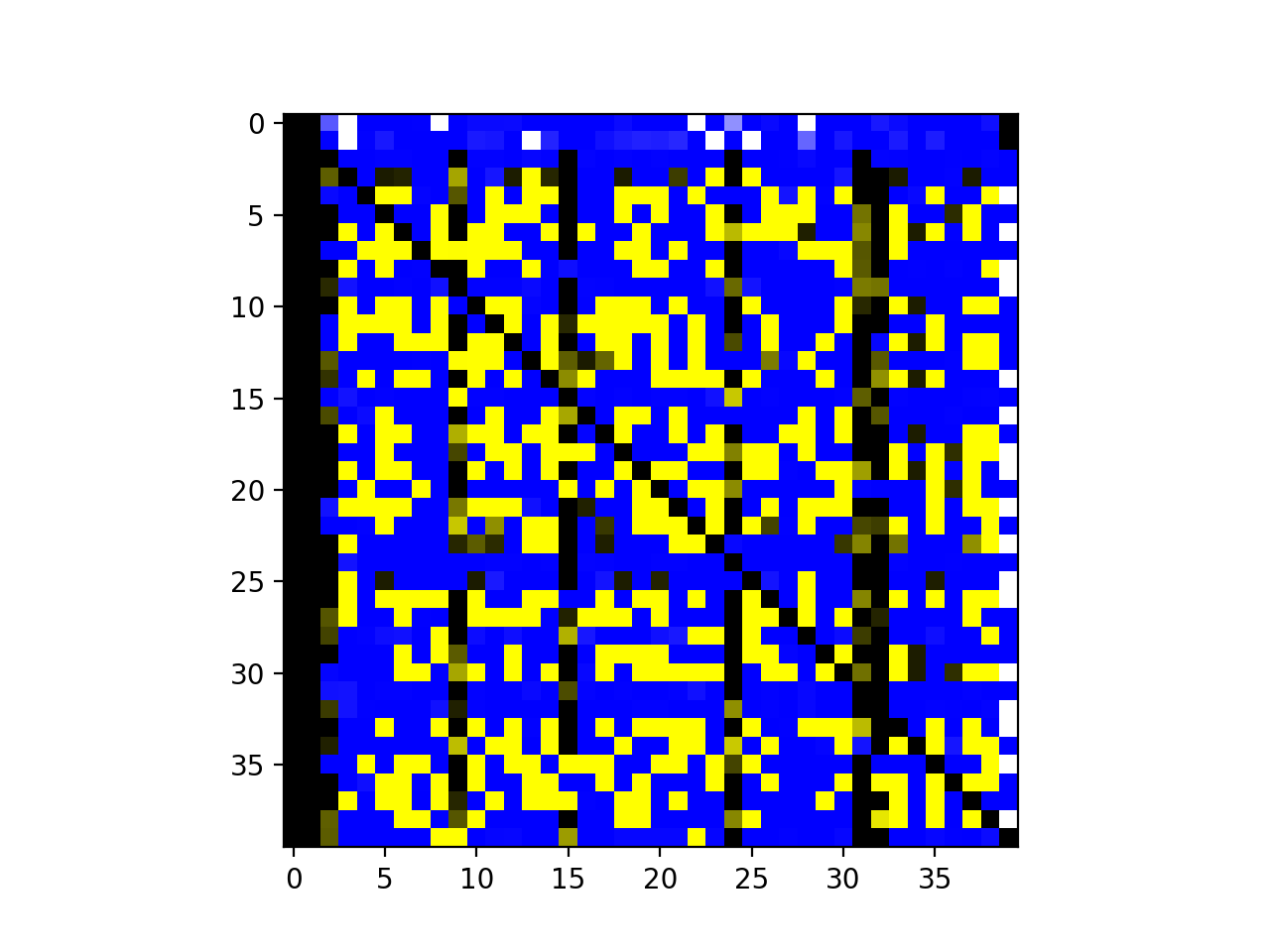} 

\caption{
\label{fig:xor}
The directed connection strengths of each of 40 neurons to each other in XOR, of which there are two input neurons and one output neuron. Yellow indicates fixed connections (i.e. those which have been rendered non-plastic by long-term memory).}
\end{figure}

We demonstrated that the architecture is able to achieve an accuracy on the task of approximately $85\%$\footnote{$\sigma^2 = 0.00676$, $n = 15$}, indicating that it was reliably solving the task with only a reward signal of $+1$ for a correct answer and $0$ otherwise. The connection strength matrix is shown in Figure~\ref{fig:xor}.

\subsection{Chrome Dino Game}
\begin{figure}
\centering
\includegraphics[width=0.6\columnwidth, height=0.5\columnwidth]{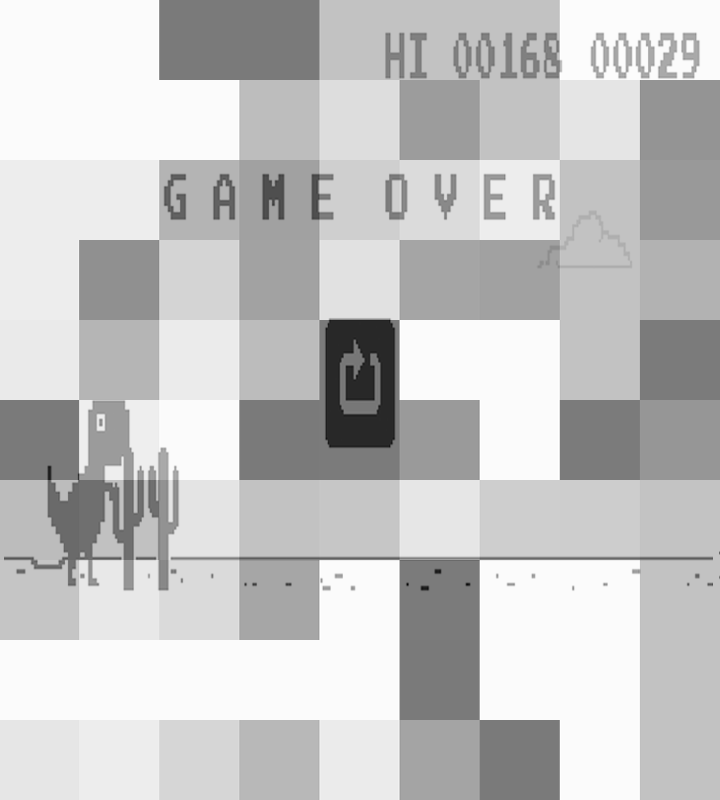} 

\caption{
\label{fig:neuron}
A pixel-by-pixel representation of a randomly selected neurons' dependencies on the visual inputs. Darker colors reduce its likelihood of firing, while brighter colors increase the likelihood. Here, we did not find a compelling pattern to explain the neuron's sensitivity.}
\end{figure}

For the behavioral learning portion, we extended our visual input stream to a screen-captured screen of the dinosaur game that Chrome displays when it is offline\footnote{Which can be accessed by going to \url{chrome://dino} in the Google Chrome browser}, and modulated the amount of learning that occurred as a function of whether or not the screen was changing. While the ultimate goal is to find an intrinsic reward function that allows the modulation to learn to play arbitrary games well while seeking novelty, this lets us test whether this approach to behavior optimization is capable of solving reinforcement learning tasks. We show what the neuron architecture sees in Figure~\ref{fig:neuron}.

Because the discretized STDP update rule for this neural net is simpler (both computationally and mathematically) than the traditional window-based approach \cite{Kempter1998SpikeBasedCT}, it wasn't immediately clear that this would have any success at all. On the dinosaur task, the best achieved performance was a score of 168. In comparison, the average score of an agent performing at chance (i.e. continuously holding down the jump key) was 57, with a maximum score of 100\footnote{$n=20$}.
Here is a recording of a typical run: \url{https://giant.gfycat.com/EmotionalSecondhandHagfish.webm}.

\begin{figure}
\centering
\includegraphics[width=0.6 \columnwidth]{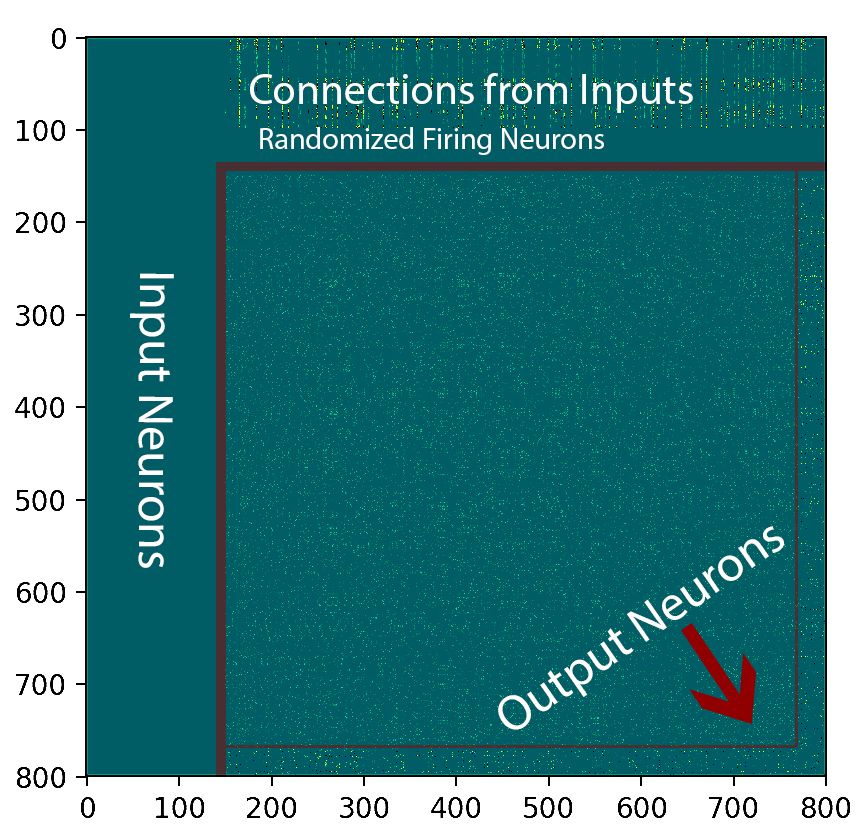}

\caption{
\label{fig:confusion}
Confusion matrix of connection strengths between 800 neurons after ten thousand frames of the dinosaur game. The left side corresponds to the neurons from which the connection originates (presynaptic), and the right side is the neuron it feeds into (postsynaptic).}
\end{figure}

The confusion matrix that results from ten thousand learning steps is shown in Figure~\ref{fig:confusion}. The neural configuration corresponds to a 10x10 pixel grayscale input, 50 randomly firing neurons, 600 "hidden" neurons, and 50 output neurons. Notably, the number of connections made by each input neuron suggest that the localization of the initial connections is failing. 

\vfill
\subsection{Mountain Car}
Here, we test our model on a classic reinforcement learning problem available in OpenAI Gym. This particular environment has a discrete action space: go left, stay still, go right. We use the observations returned by the environment as input into our neuron architecture, and we use a measure of novelty to be the reward signal that updates our model. The proportion of neurons that are firing determines the action and are discretized to fit the discrete action space. 

\begin{figure}
\centering
\includegraphics[width=1.1\columnwidth]{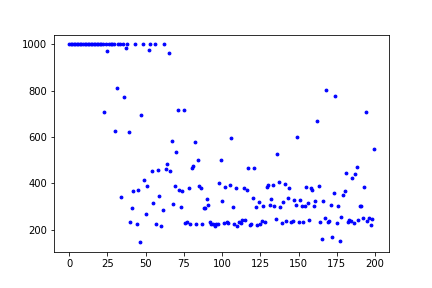} 

\caption{
\label{fig:discrete}
\textit{Discrete Mountain Car Results.} The y-axis represents the number of steps the agent needed to reach the goal. There is a limit set at 1000 steps before the episode automatically terminates. The x-axis represents the number of episodes for which the agent has been learning.}
\end{figure}

The results of the experiment indicate that the agent is able to learn to quickly reach the goal state using a measure of novelty as an instantaneous reward. We present the results of the experiment in Figure \ref{fig:discrete}.

We can see that the agent is unable to complete the objective at the beginning and is able to quickly learn the correct back-and-forth movement to reach the goal. The performance is stochastic, but the overall trend of improvement is clear. After around 75 episodes, the agent is able to reach the goal in 200 steps relatively consistently. This is not considered a successful solve of the environment, but our agent is not using the reward of the environment to learn, so this may be the agent's best performance with the novelty metric.

\vfill
\subsection{Mountain Car Continuous}
\begin{figure}
\centering
\includegraphics[width=1.1\columnwidth]{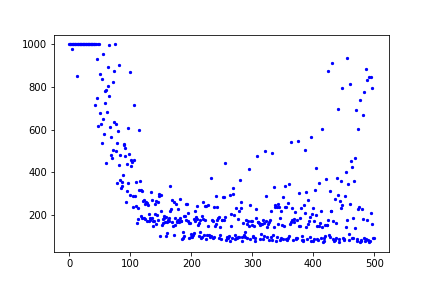} 

\caption{
\label{fig:continuous}
\textit{Continuous Mountain Car Results}. The y-axis represents the number of steps the agent needed to reach the goal. There is a limit set at 1000 steps before the episode automatically terminates. The x-axis represents the number of episodes the agent has been learning for.}
\end{figure}

Here, we adapt our model to a similar reinforcement learning problem available in OpenAI Gym, the continuous action space version of Mountain Car. This particular environment accepts any float value as the action with negative corresponding to left and positive corresponding to right (and magnitude corresponding to the impulse level). Again, we use the observations returned by the environment as input into our neuron architecture, and we use a measure of novelty to be the reward signal that updates our model. The proportion of neurons that are firing determines the action and are linearly scaled to fit the continuous action space. We present the results of the experiment in Figure~\ref{fig:continuous}.

We can see that the agent is unable to complete the objective at the beginning and is able to gradually learn the correct back-and-forth movement to reach the goal. At around 300 episodes, the agent is able to reach the goal in 100 steps or less, which is considered a successful solve of this environment.

\subsection{Pendulum}

The OpenAI Gym Inverted Pendulum environment is a classic reinforcement learning problem, though a notoriously difficult one to solve. The input was the observation of angle and angular velocity of the pendulum, and the reward here was the continuous reward returned by the environment. 

\begin{figure}[H]
\centering
\includegraphics[width=\columnwidth]{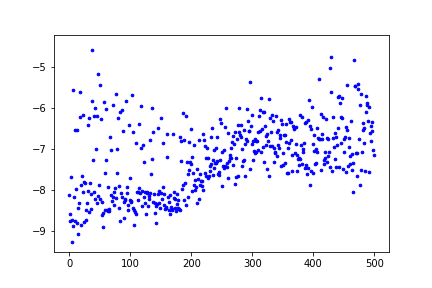} 

\caption{
\label{fig:pendulum}
\textit{Pendulum Results.} The y-axis represents the average reward per step for an episode. The x-axis represents the number of episodes for which the agent has been learning. There is a limit set at 1000 steps before the episode automatically terminates.}
\end{figure}

The performance is stochastic, but the general trend of improvement is clear. Over time, the agent learns to improve its average reward per step. The best reward at a step is 0, and the worst reward is around -16. Each episode runs for 1000 time steps. We present the results in Figure~\ref{fig:pendulum}. 

\section{Conclusion}
We experimented with neuroscientific algorithms to approach a variety of problems including pattern differentiation, Chrome Dino, XOR, and several classic reinforcement learning problems, including Mountain Car (continuous and discrete) and Pendulum. Our approach implemented spike-timing dependent plasticity, the formation of long and short-term memory, and an intrinsic curiosity-driven reward signal. Our project successfully created a generalizable agent that is capable of learning to solve problems with or without being given an explicit reward signal. Overall, we demonstrate a unique, neuro-inspired approach to reinforcement learning.

Future work would include developing a mode of learning with lower variance and less stochasticity so that performance is more consistent. Ideally, performance improvement would be relatively monotonic with slight fluctuations due to exploration. We also aim to extend our model to a wider range of reinforcement learning problems, especially an environment like MuJoCo where our novelty metric could be more useful to the agent. We also aim to implement more biologically-inspired mechanisms in our algorithm to more closely approximate the brain.

\bibliographystyle{icml2018}
\bibliography{references}

\begin{thebibliography}{17}
\providecommand{\natexlab}[1]{#1}
\providecommand{\url}[1]{\texttt{#1}}
\expandafter\ifx\csname urlstyle\endcsname\relax
  \providecommand{\doi}[1]{doi: #1}\else
  \providecommand{\doi}{doi: \begingroup \urlstyle{rm}\Url}\fi

\bibitem[Debanne et~al.(1994)Debanne, G{\"a}hwiler, and
  Thompson]{debanne1994asynchronous}
Debanne, D., G{\"a}hwiler, B., and Thompson, S.~M.
\newblock Asynchronous pre-and postsynaptic activity induces associative
  long-term depression in area ca1 of the rat hippocampus in vitro.
\newblock \emph{Proceedings of the National Academy of Sciences}, 91\penalty0
  (3):\penalty0 1148--1152, 1994.

\bibitem[Donahoe(1997)]{DONAHOE1997336}
Donahoe, J.~W.
\newblock Chapter 18 - selection networks: Simulation of plasticity through
  reinforcement learning.
\newblock In Donahoe, J.~W. and Dorsel, V.~P. (eds.), \emph{Neural-Network
  Models of Cognition}, volume 121 of \emph{Advances in Psychology}, pp.\  336
  -- 357. North-Holland, 1997.
\newblock \doi{https://doi.org/10.1016/S0166-4115(97)80104-5}.
\newblock URL
  \url{http://www.sciencedirect.com/science/article/pii/S0166411597801045}.

\bibitem[Engel et~al.(2015)Engel, Chaisangmongkon, Freedman, and
  Wang]{Engel2015}
Engel, T.~A., Chaisangmongkon, W., Freedman, D.~J., and Wang, X.-J.
\newblock Choice-correlated activity fluctuations underlie learning of neuronal
  category representation.
\newblock \emph{Nature Communications}, 6:\penalty0 6454 EP --, Mar 2015.
\newblock URL \url{https://doi.org/10.1038/ncomms7454}.
\newblock Article.

\bibitem[{Florian}(2005)]{1595864}
{Florian}, R.~V.
\newblock A reinforcement learning algorithm for spiking neural networks.
\newblock In \emph{Seventh International Symposium on Symbolic and Numeric
  Algorithms for Scientific Computing (SYNASC'05)}, pp.\  8 pp.--, Sep. 2005.
\newblock \doi{10.1109/SYNASC.2005.13}.

\bibitem[Florian(2007)]{florian2007reinforcement}
Florian, R.~V.
\newblock Reinforcement learning through modulation of spike-timing-dependent
  synaptic plasticity.
\newblock \emph{Neural Computation}, 19\penalty0 (6):\penalty0 1468--1502,
  2007.

\bibitem[{Haber} et~al.(2018){Haber}, {Mrowca}, {Fei-Fei}, and
  {Yamins}]{2018arXiv180207442H}
{Haber}, N., {Mrowca}, D., {Fei-Fei}, L., and {Yamins}, D.~L.~K.
\newblock {Learning to Play with Intrinsically-Motivated Self-Aware Agents}.
\newblock \emph{arXiv e-prints}, February 2018.

\bibitem[Hiratani \& Fukai(2014)Hiratani and Fukai]{pmid25007209}
Hiratani, N. and Fukai, T.
\newblock {{I}nterplay between short- and long-term plasticity in cell-assembly
  formation}.
\newblock \emph{PLoS ONE}, 9\penalty0 (7):\penalty0 e101535, 2014.

\bibitem[Kasabov(2014)]{kasabov2014neucube}
Kasabov, N.~K.
\newblock Neucube: A spiking neural network architecture for mapping, learning
  and understanding of spatio-temporal brain data.
\newblock \emph{Neural Networks}, 52:\penalty0 62--76, 2014.

\bibitem[Kempter et~al.(1998)Kempter, Gerstner, and van
  Hemmen]{Kempter1998SpikeBasedCT}
Kempter, R., Gerstner, W., and van Hemmen, J.~L.
\newblock Spike-based compared to rate-based hebbian learning.
\newblock In \emph{NIPS}, 1998.

\bibitem[Lee et~al.(2018)Lee, Panda, Srinivasan, and
  Roy]{10.3389/fnins.2018.00435}
Lee, C., Panda, P., Srinivasan, G., and Roy, K.
\newblock Training deep spiking convolutional neural networks with stdp-based
  unsupervised pre-training followed by supervised fine-tuning.
\newblock \emph{Frontiers in Neuroscience}, 12:\penalty0 435, 2018.
\newblock ISSN 1662-453X.
\newblock \doi{10.3389/fnins.2018.00435}.
\newblock URL
  \url{https://www.frontiersin.org/article/10.3389/fnins.2018.00435}.

\bibitem[Longtin(2013)]{Longtin:2013}
Longtin, A.
\newblock {N}euronal noise.
\newblock \emph{Scholarpedia}, 8\penalty0 (9):\penalty0 1618, 2013.
\newblock \doi{10.4249/scholarpedia.1618}.
\newblock revision \#137114.

\bibitem[Malik et~al.(2013)Malik, Gillespie, and
  Hodge]{10.3389/fncir.2013.00052}
Malik, B., Gillespie, J., and Hodge, J.
\newblock Cask and camkii function in the mushroom body α′/β′ neurons
  during drosophila memory formation.
\newblock \emph{Frontiers in Neural Circuits}, 7:\penalty0 52, 2013.
\newblock URL
  \url{https://www.frontiersin.org/article/10.3389/fncir.2013.00052}.

\bibitem[Martinez-Rubio et~al.(2018)Martinez-Rubio, Paulk, McDonald, Widge, and
  Eskandar]{Martinez-Rubio1942}
Martinez-Rubio, C., Paulk, A.~C., McDonald, E.~J., Widge, A.~S., and Eskandar,
  E.~N.
\newblock Multimodal encoding of novelty, reward, and learning in the primate
  nucleus basalis of meynert.
\newblock \emph{Journal of Neuroscience}, 38\penalty0 (8):\penalty0 1942--1958,
  2018.
\newblock ISSN 0270-6474.
\newblock \doi{10.1523/JNEUROSCI.2021-17.2017}.
\newblock URL \url{http://www.jneurosci.org/content/38/8/1942}.

\bibitem[Rescorla \& Wagner(1972)Rescorla and Wagner]{rescorla1972theory}
Rescorla, R.~A. and Wagner, A.~R.
\newblock A theory of pavlovian conditioning: Variations in the effectiveness
  of reinforcement and nonreinforcement.
\newblock \emph{Classical conditioning II: Current research and theory}, pp.\
  64--99, 1972.

\bibitem[Salgado et~al.(2012)Salgado, K\"{o}hr, and Trevi{\~{n}}o]{Salgado2012}
Salgado, H., K\"{o}hr, G., and Trevi{\~{n}}o, M.
\newblock Noradrenergic `tone' determines dichotomous control of cortical
  spike-timing-dependent plasticity.
\newblock \emph{Scientific Reports}, 2\penalty0 (1), may 2012.
\newblock \doi{10.1038/srep00417}.
\newblock URL \url{https://doi.org/10.1038/srep00417}.

\bibitem[Stanley \& Miikkulainen(2002)Stanley and
  Miikkulainen]{Stanley:2002:ERL:2955491.2955578}
Stanley, K.~O. and Miikkulainen, R.
\newblock Efficient reinforcement learning through evolving neural network
  topologies.
\newblock In \emph{Proceedings of the 4th Annual Conference on Genetic and
  Evolutionary Computation}, GECCO'02, pp.\  569--577, San Francisco, CA, USA,
  2002. Morgan Kaufmann Publishers Inc.
\newblock ISBN 1-55860-878-8.
\newblock URL \url{http://dl.acm.org/citation.cfm?id=2955491.2955578}.

\bibitem[Wise \& Rompre(1989)Wise and Rompre]{wise1989brain}
Wise, R.~A. and Rompre, P.-P.
\newblock Brain dopamine and reward.
\newblock \emph{Annual review of psychology}, 40\penalty0 (1):\penalty0
  191--225, 1989.

\end{thebibliography}


\end{document}